\documentclass[letterpaper, 10 pt, conference]{ieeeconf}  
                                                          
\IEEEoverridecommandlockouts                              
                                                          
\usepackage{cite}
\usepackage{amsmath,amssymb,amsfonts}
\usepackage{algorithmic}
\usepackage{graphicx}
\graphicspath{{./figures/}}
\usepackage{caption}
\usepackage{subcaption}
\usepackage{textcomp}
\usepackage{xcolor}
\usepackage{todonotes}
\usepackage{tabularx,booktabs}
\usepackage{multirow}

\DeclareMathOperator*{\argmin}{arg\,min}

\title{Pole-based Vehicle Localization with Vector Maps:\\A Camera-LiDAR Comparative Study}
\author{
Maxime Noizet$^1$
\and
Philippe Xu$^1$
\and
Philippe Bonnifait$^1$
\thanks{$^{1}$The authors are with the Universit\'e de technologie de Compi\`egne, CNRS, Heudiasyc, France.
{\tt\small name.surname@hds.utc.fr}}%
}
\begin{document}

\maketitle

\begin{abstract}

For autonomous navigation, accurate localization with respect to a map is needed. 
In urban environments, infrastructure such as buildings or bridges cause major difficulties to Global Navigation Satellite Systems (GNSS) and, despite advances in inertial navigation, it is necessary to support them with other sources of exteroceptive information.
In road environments, many common furniture such as traffic signs, traffic lights and street lights take the form of poles.
By geo-referencing these features in vector maps, they can be used within a localization filter that includes a detection pipeline and a data association method. 
Poles, having discriminative vertical structures, can be extracted from 3D geometric information using LiDAR sensors. Alternatively, deep neural networks can be employed to detect them from monocular cameras. The lack of depth information induces challenges in associating camera detections with map features.
Yet, multi-camera integration provides a cost-efficient solution.
This paper quantitatively evaluates the efficacy of these approaches in terms of localization.
It introduces a real-time method for camera-based pole detection using a lightweight neural network trained on automatically annotated images. 
The proposed methods' efficiency is assessed on a challenging sequence with a vector map. 
The results highlight the high accuracy of the vision-based approach in open road conditions.

\end{abstract}

\section{Introduction}
\label{sec:intro}

In the field of autonomous driving, achieving a reliable and accurate localization solution is crucial to ensure safe and efficient navigation when using a navigation map. For example, localization is essential for tasks such as planning, crossing intersections, aiding perception, cooperative navigation, etc. Depending on the context and the requirements of the localization task, this can be particularly challenging. Even on rather favourable operational domains like highways, non-differential multi-constellation GNSS (Global Navigation  Satellite Systems) aided by Dead-Reckoning (DR) sensors is insufficient when lane-level positioning is needed \cite{laconte_survey_2021}.



To improve the localization performance, exteroceptive sensors such as LiDARs or cameras can be added to handle the mentioned limitations in complex environments. 
In this case, a vector map can be an efficient and scalable means to manage geo-referenced features such as traffic signs, lane markings or other road features. In this paper, we focus on High-Definition vector maps (HD maps) with a cm-level accuracy. 

To reach lane-level positioning, lane markings and curbs are now very well detected by cameras and associated with HD maps to improve cross-track accuracy and integrity \cite{frisch, al_hage_2019}. Yet, road information for localization is sensitive to environmental factors as degradation, occlusion, and variation which can lead to unreliable and inconsistent measurements. Besides, it requires an HD map containing a geometric, e.g., polyline-based, description of all road markings, which can be costly to produce and maintain.

There are other widespread road infrastructure elements that can be used as additional sources of information as they are quite easy to detect. For example, using a LiDAR point cloud, traffic signs can be easily  extracted with intensity filtering and used to improve localization\cite{ghallabi_lidar-based_2019}. 
Traffic signs represent only a small part of all the features available in a road environment. They belong to a broader widespread class which is that of poles or vertical signage including in addition traffic lights and streetlights. They can provide absolute localization information when detected by on-board sensors. They have shown to improve deeply localization performance~\cite{li_robust_2021, spangenberg_pole-based_2016}. 

\begin{figure}
    \centering
    \includegraphics[width=\columnwidth]{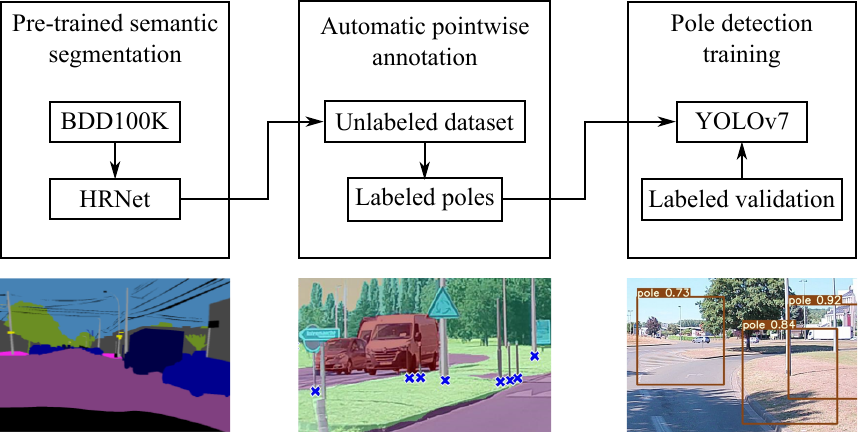}
    \caption{Image-based poles dectector training. A pre-trained semantic segmentation network is used to annotate an unlabeled custom dataset. The pixel-wise annotations are transformed into pointwise labels at the bases of the poles. A YOLOv7 is trained using bounding boxes centered at the pointwise labels. A small amount of manually labeled data are used for validation.}
    \label{fig:diag_cam}
\end{figure}

However, LiDAR sensors have some limitations due to the sparse nature of point clouds. Detecting fine or distant structures can be complex, and the cost of such a sensor can also be a barrier for large-scale deployment. In contrast, monocular cameras do not have these drawbacks, although their field of view is necessarily narrower and they are more sensitive to illumination conditions. 

Many methods can be applied to detect objects in images captured by cameras. To the best of our knowledge, there is limited research on pole detection in images\cite{sharma_image_2015,zhang_using_2018}, particularly from a localization perspective \cite{Barbosa2021SoftCA}.
In a previous work \cite{IV23}, we proposed a pole bases detector using a neural network trained on automatically annotated data using HD-maps. 

Monocular cameras only provide angular information resulting in a bearing-only localization problem. Bearing information has been used in various studies as in tracking applications in aviation or submarine fields, but also in vehicle localization. Bearing-only Simultaneous Localization And Mapping (SLAM) have been studied\cite{bekris_evaluation_2006}. Some camera-based methods \cite{jensfelt_framework_2006,lemaire_vision-based_2007} have been proposed where the visual features used are low-level features such as SIFT\cite{lowe1999object} or Harris-Laplace points\cite{Harris1988ACC}.
In \cite{omni_cam_landmark_arrangement}, an omni-camera and four landmarks are used to localize an automated agricultural vehicle with distinguishable landmarks optimally placed. 

Pole-based localization is very challenging for data association due to the non-discernability of road features. Incorrect associations of pole detections with vector map features can therefore lead to poor localization. Moreover, relying solely on bearing information makes it difficult to accurately estimate a vehicle's pose since angular measurements are made relatively to the heading of the vehicle.

In this paper, the objective is to study how a localization system based on GNSS and DR sensors can be improved by incorporating pole-like feature detections that are matched with a vector map. We consider detections obtained from both a LiDAR and a multi-camera system, and compare their performance in terms of accuracy improvements. 

The article is organized as follows. 
In Section \ref{sec:detection}, we propose a monocular image-based pole detection method using a semantic segmentation network to generate pseudo labels to train an object detector. The LiDAR-based geometric pole detector is also described. The pole-based localization framework is given in Section  \ref{sec:localization}. Finally, experimental results using real data are detailed and analyzed in Section \ref{sec:results}.
Finally, Section \ref{sec:conclusion} presents conclusions and future work.


\section{Pole detection}
\label{sec:detection}
\subsection{Problem statement}

The thickness of pole-like features with respect to the scale of an urban HD vector map implies that they are usually mapped as points.
The coordinates of such a point typically represent the base of a pole at the ground level.
In the rest of the paper, we consider the general case where the poles are mapped as 2D points without further information about their types, their height or geometry.
From the map perspective, all the poles are considered as being indistinguishable.
The vector map is therefore a set of georeferenced landmarks
\begin{align}
    \boldsymbol{\mathcal{M}}=\left\{\left.{}^{(\text{O})}\boldsymbol{m}_j\in\mathbb{R}^2\right|j=1,\ldots \right\}
\end{align}
where each map feature ${}^{(\text{O})}\boldsymbol{m}_j$ is a 2D point expressed in a local working frame \((\text{O})\) using East-North-Up (ENU) coordinates.
For clarity, in the rest of the paper, the left exponent will be omitted when the coordinates are expressed in the \((\text{O})\) frame.

The aim is to detect these features using cameras and LiDAR.
In the image frame, it consists in detecting the pixel coordinates of the bases of the poles.
In the LiDAR point cloud, it comes to compute the 2D coordinates of the projections of the poles onto the ground plane.

\subsection{Camera-based detection}

The visual characteristics of poles make them ill suited to be detected from an object-based bounding box point of view.
Indeed, poles are thin and are often truncated in the image field-of-view.
Therefore, poles are most of the time considered at the pixel-level within a semantic segmentation framework.
The main drawback of dense semantic segmentation is its computational cost compared to modern object detection such as YOLO~\cite{redmon2016}.

In our prior work~\cite{IV23}, we have demonstrated how to formalize the detection of poles in images with an object detection pipeline using bounding boxes centered at the bases of the poles, \textit{i.e.}, the contact point between the poles and the ground.
The labels for the images were automatically generated by the joint use of a vector map and a LiDAR in order to compute the projection of the map features onto the image frame.
For this purpose, it was necessary to estimate the ground plane as well as determining whether or not the feature was visible, \textit{e.g.}, not occluded by some obstacles.
One of the limitations of this solution is that it requires a LiDAR in addition to the cameras as well as a localization ground truth.

We propose to extend the solution in~\cite{IV23} by making use of a pre-trained semantic segmentation neural network to generate pointwise annotation.
The process follows the steps pictured in Fig.~\ref{fig:diag_cam}:
\begin{enumerate}
    \item Train a semantic segmentation neural network on a labeled dataset that includes pole-like classes;
    \item Use the neural network to generate pixel-wise pseudo-labels in an unlabeled dataset and compute point-wise labels at the bases of the poles;
    \item Use bounding boxes centered at the poles to train an object detector using a small dataset of manually annotated images for validation.
\end{enumerate}

For the first step, we use the High-Resolution Network (HRNet) proposed in~\cite{segmenter} with a model pre-trained on the BDD100K dataset~\cite{bdd100k}.
Among multiple networks trained on BDD100K dataset, it is one of the most effective for segmenting pixels related to pole-like classes. 
In this dataset, the pole-like features correspond to three classes, namely “pole”, “traffic sign” and “traffic light”.

For the second step, all connected pole-like pixels are grouped and for such a cluster the lowest pixel lying on ground pixels, if it exists, is considered as the pole base and is labeled as such.
Contrary to what has been proposed in~\cite{IV23} the pixel semantic labels are not from the ground truth but are pseudo-labels computed from a neural network.

\begin{figure*}[t!]
    \centering
    \begin{subfigure}[b]{0.325\textwidth}
        \includegraphics[width=\columnwidth]{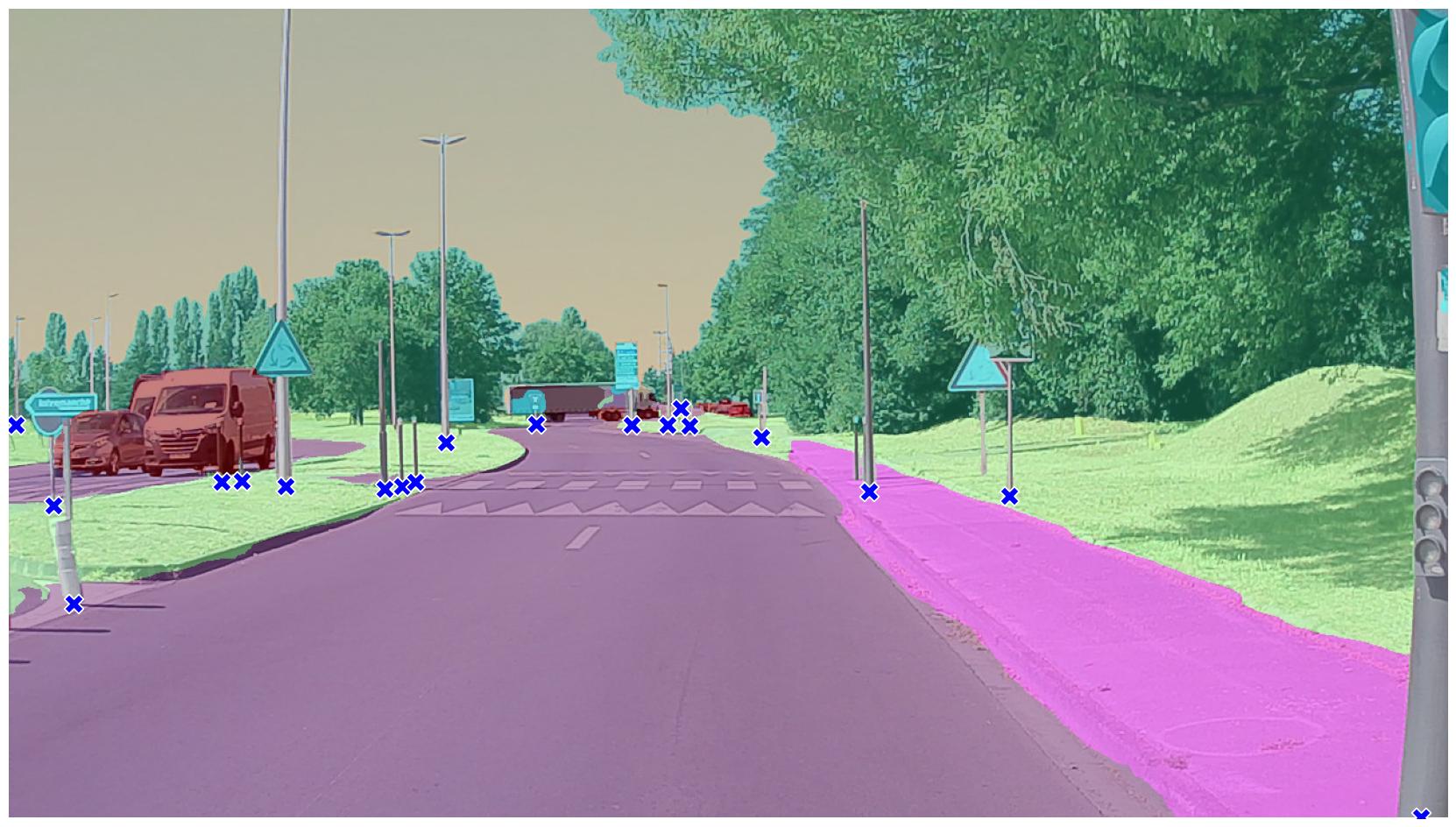}
        \caption{Front RGB image}
    \end{subfigure}
    \begin{subfigure}[b]{0.325\textwidth}
        \includegraphics[width=\columnwidth]{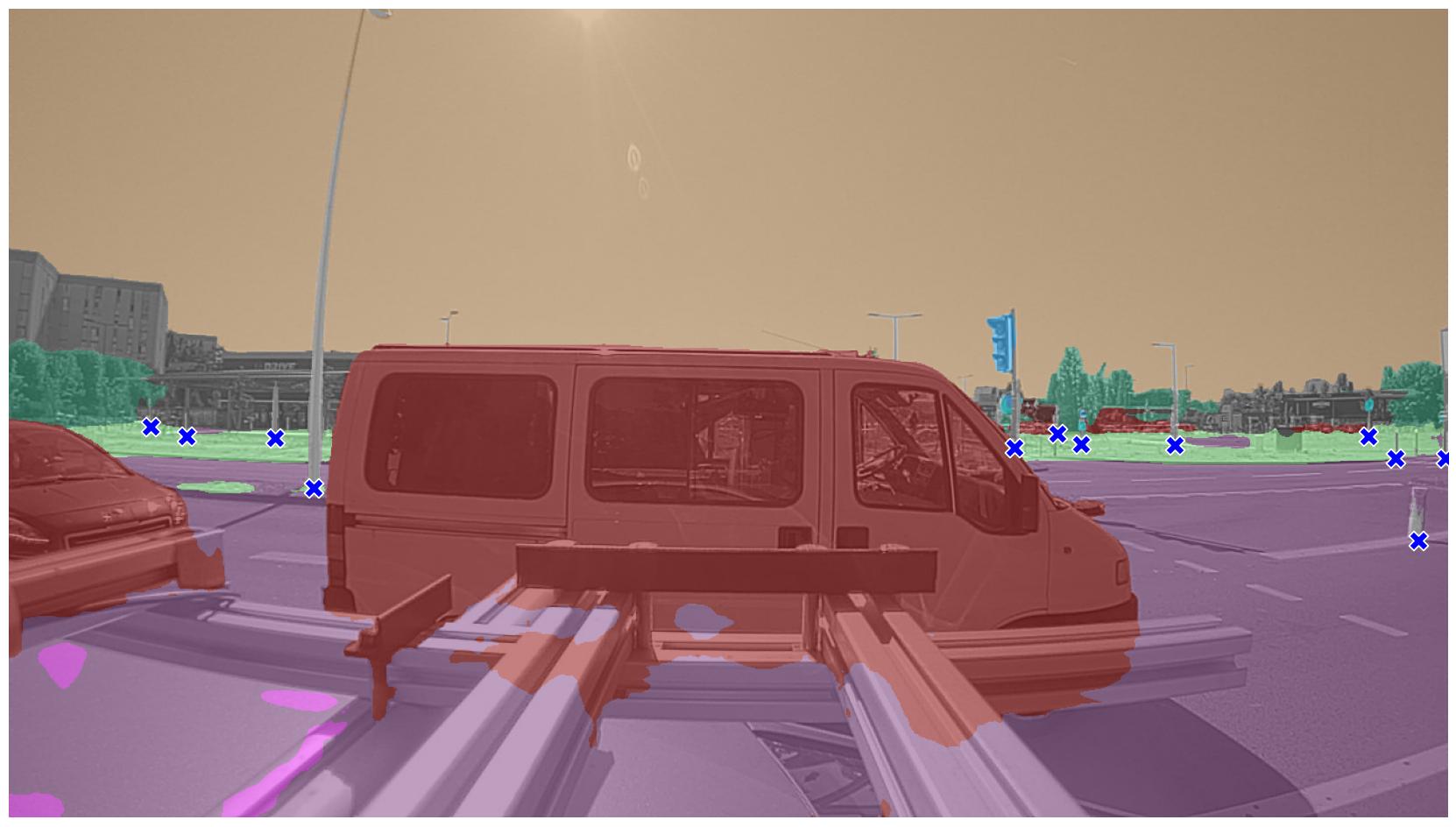}
        \caption{Left grayscale image}
    \end{subfigure}
    \begin{subfigure}[b]{0.325\textwidth}
        \includegraphics[width=\columnwidth]{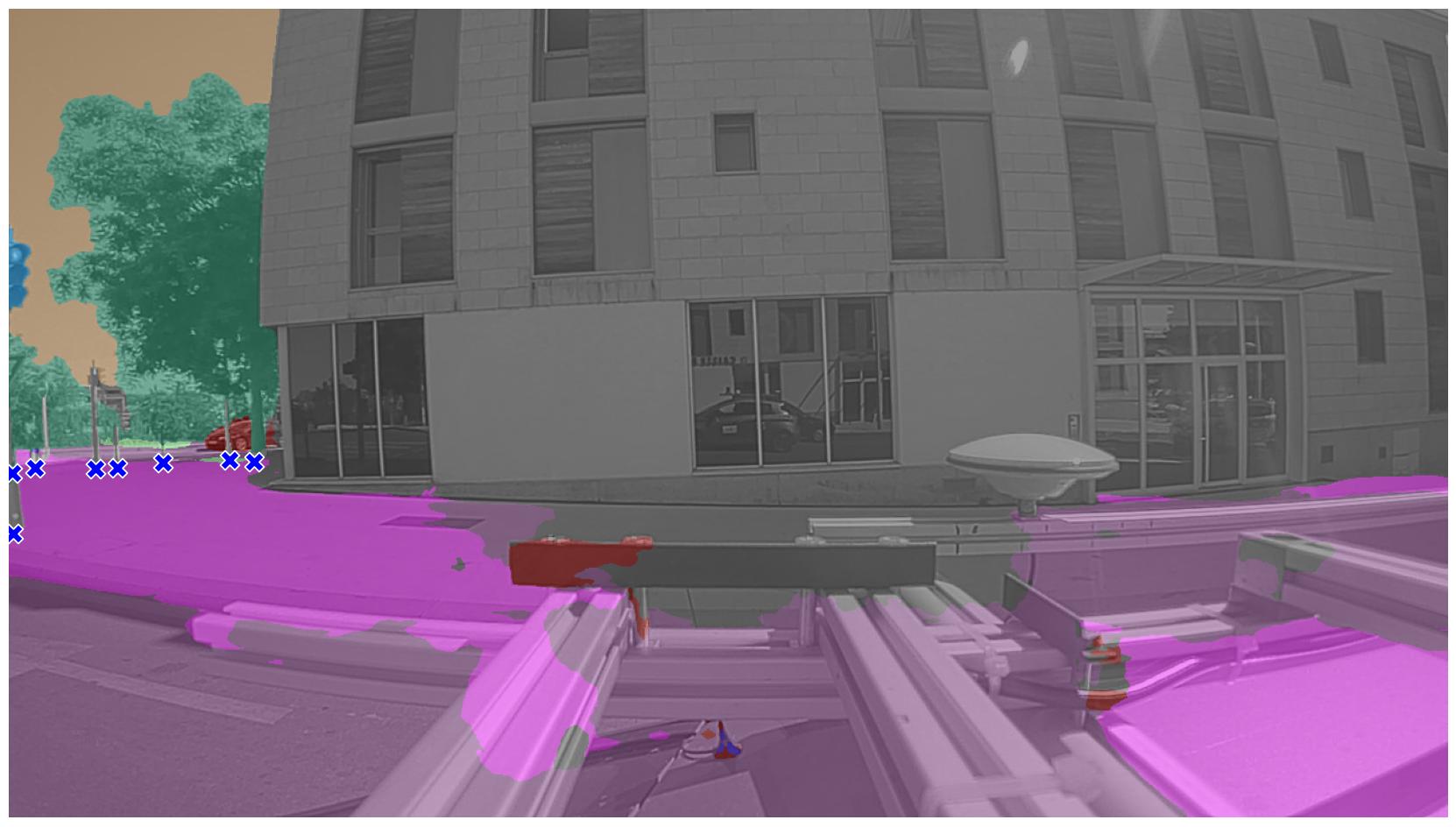}
        \caption{Right grayscale image}
    \end{subfigure}
    \caption{Examples of segmented images and the obtained annotations (blue crosses).}
    \label{fig:segmentation_annotation}
\end{figure*}

\begin{figure*}[t!]
    \centering
    \begin{subfigure}[b]{0.325\textwidth}
        \includegraphics[width=\columnwidth]{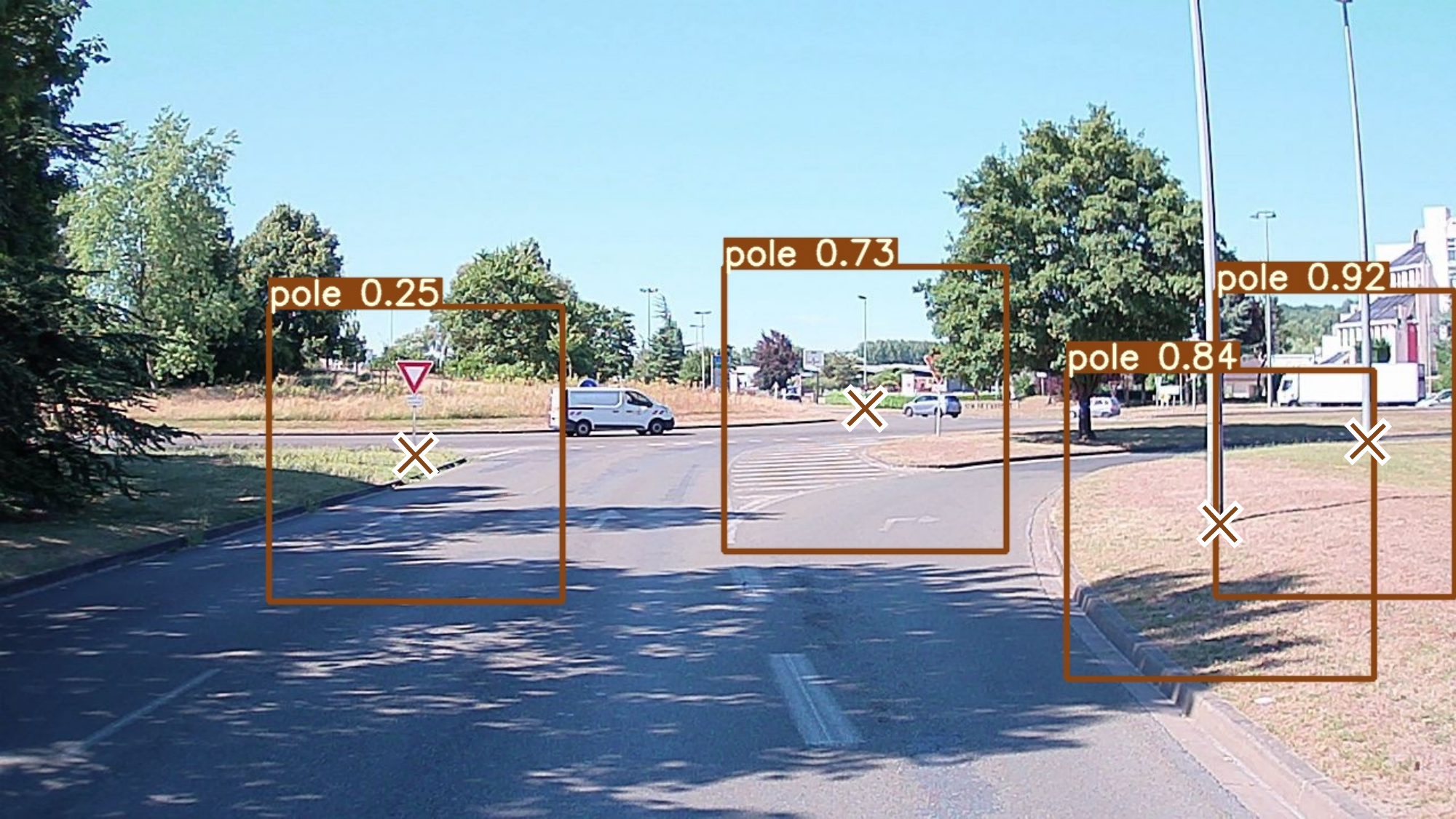}
    \end{subfigure}
    \begin{subfigure}[b]{0.325\textwidth}
        \includegraphics[width=\columnwidth]{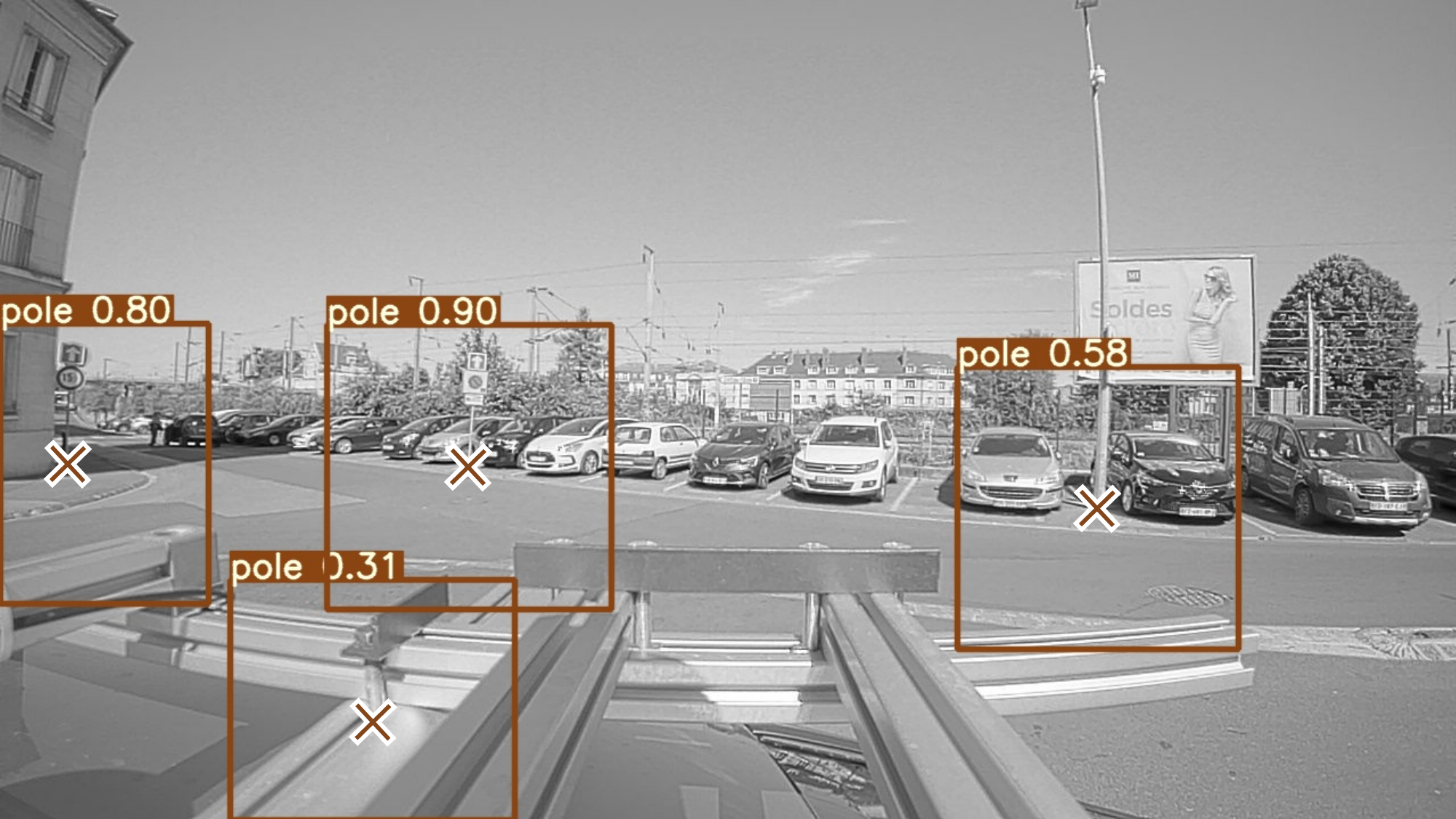}
    \end{subfigure}
    \begin{subfigure}[b]{0.325\textwidth}
        \includegraphics[width=\columnwidth]{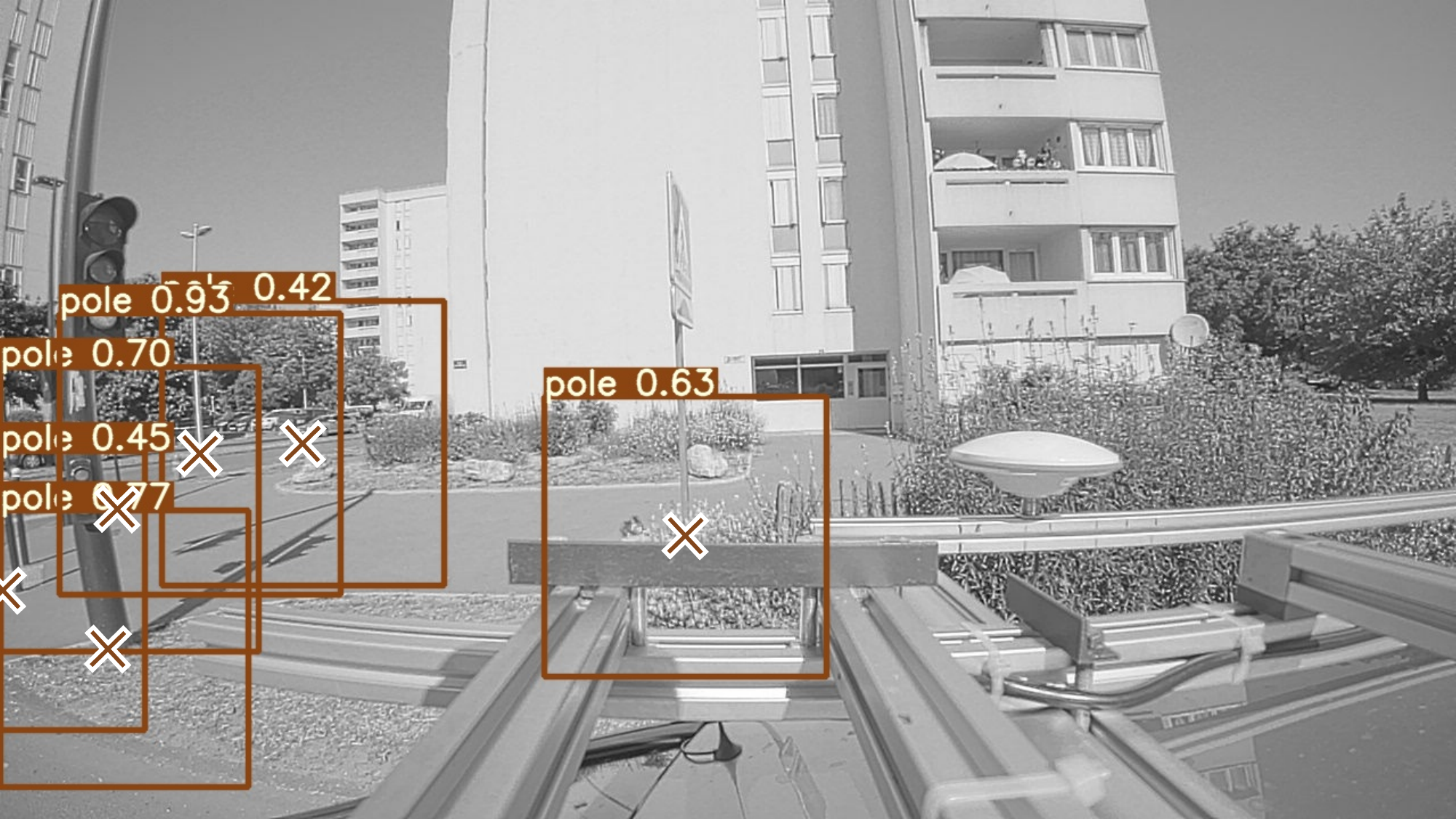}
    \end{subfigure}
    \caption{Examples of detections obtained from YOLOv7-based pole detectors on RGB and grayscale images. Each bounding box is displayed with its detection score and its center corresponding to a pole base is highlighted by a cross. For the grayscale side cameras, a final filtering is applied to remove detections on the vehicle roof.}
    \label{fig:yolo_results}
\end{figure*}

Finally, for the last step, we transform the pointwise labels into squared fixed-sized bounding boxes centered on the labeled points. 
We then feed these data to a YOLOv7~\cite{wang22} object detector as in~\cite{IV23} while tuning the size of the bounding boxes on a validation set containing a small amount manually annotated images.
At the inference stage, the detected bounding boxes are converted back into points by computing the center point.
For a given image I at time \(k\), the output of the detection is a set of measurements:
\begin{align}
    {}^{(\text{I})} \boldsymbol{Y}^{\text{I}}_k = \left\{\left. {}^{(\text{I})} \boldsymbol{y}^{\text{I}}_{k,i}=\left(u_{k,i},v_{k,i}\right) \right| i=1,\ldots  \right\},
\end{align}
where each measurement $ {}^{(\text{I})} \boldsymbol{y}^{\text{I}}_{k,i}$ is the pixel coordinates \(u,v\) of a pole base expressed in the image frame \((\text{I})\) of the camera.

We apply this strategy to a multi-camera system composed of a front color camera and two wide-angle grayscale cameras directed on the sides.
Examples of segmented images with the annotations obtained are illustrated in Fig.  \ref{fig:segmentation_annotation}. 
Even though the images in the BDD100K dataset are more similar to the color camera, the performance on the wide-angle grayscale ones are reasonable.
Fig.~\ref{fig:yolo_results} pictures the detection results on the three cameras.
For the side cameras, a final filtering is applied to remove detections on the vehicle roof.


\subsection{LiDAR-based detection}

\begin{figure}
    \centering
    \includegraphics[width=\columnwidth]{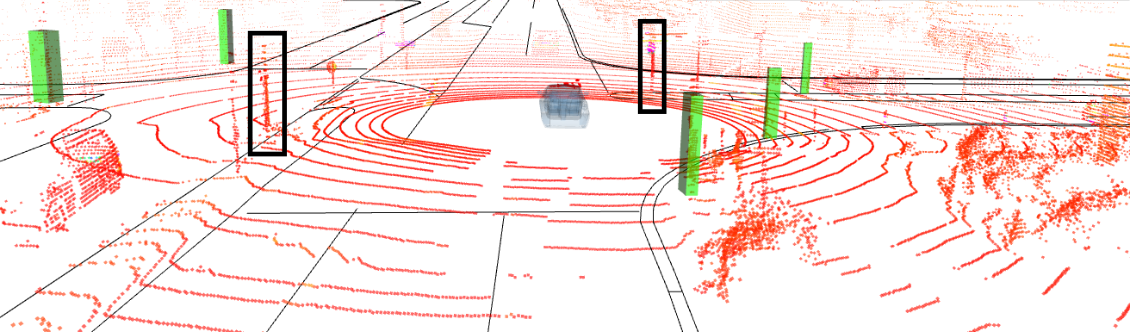}
    \caption{Examples of pole detections obtained from LiDAR. The bounding boxes of detected poles are visible in green. Two examples of missed detections are highlighted with black rectangles. Other examples are visible in point cloud distribution.}
    \label{fig:lidar_results}
\end{figure}

In a LiDAR point-cloud, each point is characterized by its Cartesian position in the vehicle frame. Consequently pole-like features can be extracted using geometric-based techniques. 
Firstly, ground points are removed and remaining points are grouped into clusters using the method proposed by Zermas et~al.~\cite{ground_seg_cluster}. Then, each obtained cluster is classified as a pole or not using a Principal Component Analysis (PCA) strategy. For each cluster, the principal components characterized by the eigenvectors $v_1,\ v_2,\ v_3$ and eigenvalues $\lambda_1,\ \lambda_2,\ \lambda_3$ of the covariance matrix sorted in descending order are computed. Then, thresholds are defined to consider a cluster as a pole:
\begin{itemize}
    \item Linearity \( l = (\lambda_1 - \lambda_2) / \lambda_1\): quantifies the predominance of the main component compared to the others. A pole should have a high linearity.
    \item Orientation \(\beta\): quantifies the angle between the \(z\)-axis and the main component direction $v_1$. A pole should be vertical, \textit{i.e.}, have a low value for \(\beta\).
    \item Height \(h\): a pole is typically a tall cluster.
    \item Thickness \(t\): a pole is typically a thin cluster.
\end{itemize}
For each of these quantities, we define a threshold and we consider a cluster as being a pole if the following condition is met:
\begin{align}
    \left( l > l_{\min} \right) \text{\&} \left( \beta < \beta_{\max} \right) \text{\&} \left( h > h_{\min} \right) \text{\&} \left( t < t_{\max} \right)
\end{align}
The detection output from the LiDAR L is a set of 2D measurements
\begin{align}
    {}^{(\text{L})}\boldsymbol{Y}^{\text{L}}_k = \left\{ \left.
    {}^{(\text{L})}\boldsymbol{y}^{\text{L}}_{k,i} = \left({}^{(\text{L})}x^{\text{L}}_{k,i},{}^{(\text{L})}y^{\text{L}}_{k,i}\right) \right| i=1,\ldots \right\},
\end{align}
where each measurement $ {}^{(\text{L})}\boldsymbol{y}^{\text{L}}_{k,i}$ corresponds to the 2D coordinates of the centroid of the cluster $i$ expressed in the LiDAR frame \((\text{L})\).
An example of poles detection using LiDAR data is illustrated in Fig. \ref{fig:lidar_results}.

\section{Pole-based localization filter}
\label{sec:localization}

We build our localization solution using a standard extended Kalman filter formalism.
The system uses GNSS, wheel speed sensors, a gyro, a LiDAR and multiple cameras for pole detection and a vector HD map.
At a time \(k\), the state vector \(\boldsymbol{x}_k\) is expressed as follows:
\begin{equation}
    \boldsymbol{x}_{k} = \left[x_k, y_k, \theta_k, v_k, \dot{\theta}_k, b_{k,x}, b_{k,y} \right]^{\top}
\end{equation}
The component \(\boldsymbol{q}_k=(x_k, y_k, \theta_k)\) is the vehicle pose, \textit{i.e.}, position and heading, defined at the center of the vehicle rear axle.
The components \(v_k\) and \(\dot{\theta}_k\) correspond to the longitudinal speed and the yaw rate, respectively.
To handle GNSS bias exhibited by the receiver, a random constant GNSS 2D bias \((b_{k,x}, b_{k,y})\) is added to the state vector.

\subsection{GNSS and DR}
The GNSS measurements \(\boldsymbol{z}^{\text{G}}_k\) provide the 2D coordinates of the antenna, its observation model is derived as follows:
\begin{align} 
    \boldsymbol{z}^{\text{G}}_k = 
    \begin{bmatrix}
        x_k + b_{k,x}\\
        y_k + b_{k,y}
    \end{bmatrix} +
    \begin{bmatrix}
        \cos\theta_k & -\sin\theta_k \\
        \sin\theta_k & \cos\theta_k 
    \end{bmatrix} 
    \begin{bmatrix}
        t_x\\
        t_y
    \end{bmatrix}
    + \boldsymbol{\beta}^\text{G}_k 
\end{align}
where \((t_x,t_y)\) is the antenna lever arm with respect to the vehicle frame and \(\boldsymbol{\beta}^{\text{G}}_k\) the GNSS observation noise.

For the dead reckoning the observation models of the left and right rear wheel speeds, \(z^{\text{W}_l}_k\) and \(z^{\text{W}_r}_k\), are given in \(\text{m}\cdot\text{s}^{-1}\) and expressed as follows:
\begin{align}
    z^{\text{W}_l}_k = v_k - \frac{\ell}{2}\dot{\theta}_k + \beta^{\text{W}_l}_k\ ,\quad
    z^{\text{W}_r}_k = v_k + \frac{\ell}{2}\dot{\theta}_k + \beta^{\text{W}_r}_k
\end{align}
where \(\ell\) the distance separating the two wheels and \(\beta^{\text{W}_l}_k\), \(\beta^{\text{W}_r}_k\) the observation noises.
Finally, the gyro provides a straightforward measurement of the yaw rate \(z^{\text{Y}}_k = \dot{\theta}_k + \beta^{\text{Y}}_k\).


\subsection{Poles measurements}

To build the observation model for the poles detected by the LiDAR or the cameras, two steps are necessary.
First, the detections from the LiDAR and the cameras need to be expressed in a common space with respect to the map features.
And second, the detections and the map features need to be associated to each other.

For the LiDAR, a detection is represented by a 2D point similar to the map representation.
We can either move the map features from the \((\text{O})\) frame to the LiDAR \((\text{L})\) frame or the opposite for solving the data-association.
Because there are often less detections than map features, the latter is less computationally demanding.

At a given time \(k\), an estimate of the vehicle pose \(\hat{\boldsymbol{q}}_{k|k-1}\) is predicted from the previous state estimate \( \hat{\boldsymbol{x}}_{k-1} \).
This pose estimate is then used to transform the LiDAR detection set \({}^{(\text{L})}\boldsymbol{Y}^{\text{L}}_k\) from the \((\text{L})\) frame to the \((\text{O})\) frame: \(\boldsymbol{Y}^{\text{L}}_k\).

For the pole detection in the image frame, because of the lack of depth information from monocular cameras, it is not possible to compute the 2D coordinates of the detected poles in the map frame.
Instead, we use a bearing only approach to encode the poles detection.
Given the camera intrinsic calibration parameters, the image detection set \({}^{(\text{I})} \boldsymbol{Y}^{\text{I}}_k\) is transformed into a set of bearing angles expressed in the camera frame \((\text{C})\):
\begin{align}
    {}^{(\text{C})} \boldsymbol{Y}^{\alpha}_k = \left\{\left. {}^{(\text{C})}y^{\alpha}_{k,i}=\alpha_{k,i}\in\left[-\pi;\pi\right)\right| i=1,\ldots  \right\}
\end{align}
where \(\alpha_{k,i}\) corresponds to the angle of the \(i\)-th detection with respect to the direction pointed by the camera which is aligned with the vehicle heading in the case of the front color camera.

Contrary to the LiDAR case, for the camera, it is the map features that are transformed into the camera frame.
Given \(\hat{\boldsymbol{q}}_{k|k-1}\), the map features within a limited radius around the pose position, are transformed into the camera frame and their relative angles with respect to the camera are computed.
This leads to a camera map composed of angles relative to the camera frame:
\begin{align}
    {}^{(\text{C})} \boldsymbol{\mathcal{M}}^{\alpha}_k=\left\{\left.{}^{(\text{C})}m^{\alpha}_{k,j}=\alpha_{k,j}\in\left[-\pi;\pi\right)\right|j=1,\ldots \right\}
\end{align}
The same process is done for each of the three cameras.







\subsection{From measurements to map-matched observations}

Map-matching consists in associating the measurements of the detected features with landmarks retrieved from the map.
Because we have considered the features to be indistinguishable, we use geometric distances to associate the data.
The Mahalanobis distance can be used to measure the proximity of a LiDAR measurement \( \boldsymbol{y}_{k,i}^{\text{L}} \) and a map feature \(\boldsymbol{m}_j\) as follows:
\begin{align}
D^\text{L}_{k,i,j}=\sqrt{(\boldsymbol{m}_j-\boldsymbol{y}^\text{L}_{k,i})^\top 
    {R^\text{L}_{k,i}}^{-1}
    (\boldsymbol{m}_j-\boldsymbol{y}^\text{L}_{k,i})^\top}    
\end{align}
where \(R^\text{L}_{k,i}\) is the covariance matrix associated to the LiDAR measurement \( \boldsymbol{y}_{k,i}^{\text{L}} \) computed from the covariance matrix of the pose estimate \(\hat{\boldsymbol{q}}_{k|k-1}\).

In the camera case, we manipulate angular quantities, for a camera measurement \({}^{(\text{C})} y^{\alpha}_{k,i}\in\left[-\pi;\pi\right)\) and a map feature \({}^{(\text{C})}m^{\alpha}_{k,j}\in\left[-\pi;\pi\right)\), their difference \(\delta_{k,i,j}\) is mapped onto the \(\left[-\pi;\pi\right)\) interval as follows:
\begin{align}
    \delta_{k,i,j} = \left[\left( {}^{(\text{C})}m^{\alpha}_{k,j} - {}^{(\text{C})} y^{\alpha}_{k,i} + \pi\right) \text{ mod } 2\pi \right] - \pi
\end{align}
where mod \(2\pi\) is the modulo operator providing the result within \(\left[0;2\pi\right)\). 
The final distance is then defined as the squared difference
\begin{align}
    D^\text{C}_{k,i,j} = \delta_{k,i,j}^2
\end{align}

The map-matching problem is solved as an assignment problem using the Hungarian method~\cite{kuhn1955hungarian}.
This method finds in a polynomial time the optimal sets of pairs
\begin{align}
    \boldsymbol{Z}^\text{L}_k &= \left\{ \boldsymbol{z}^\text{L}_{k,i,j} = \left( \boldsymbol{y}^\text{L}_{k,i},\ \boldsymbol{m}_j\right) \right\} \\
    \boldsymbol{Z}^\text{C}_k &= \left\{ \boldsymbol{z}^\text{C}_{k,i,j} = \left( {}^{(\text{C})} y^{\alpha}_{k,i},\ {}^{(\text{C})}m^{\alpha}_{k,j}\right) \right\} 
\end{align}
that minimize the sum of the associated distances 
\begin{align}
    \boldsymbol{Z}^\text{L}_k &= \argmin_{\boldsymbol{y}^\text{L}_{k,i}\in\boldsymbol{Y}^\text{L}_{k}, \boldsymbol{m}_j\in\boldsymbol{\mathcal{M}}} \sum_{i,j} D^\text{L}_{k,i,j} \\
    \boldsymbol{Z}^\text{C}_k &= \argmin_{{}^{(\text{C})} y^{\alpha}_{k,i}\in{}^{(\text{C})} \boldsymbol{Y}^{\alpha}_{k}, {}^{(\text{C})}m^{\alpha}_{k,j}\in{}^{(\text{C})} \boldsymbol{\mathcal{M}}^{\alpha}_k} \sum_{i,j} D^\text{C}_{k,i,j}
\end{align}
under the constraint that at most one measurement can be associated to a map feature and conversely.

Once the map-matching step is done, the observations from the LiDAR and the cameras are injected into the localization filter.

\section{Experimental results}
\label{sec:results}

\subsection{Experimental setup}

\begin{figure}[t!]
 \centering
 \includegraphics[width=\columnwidth]{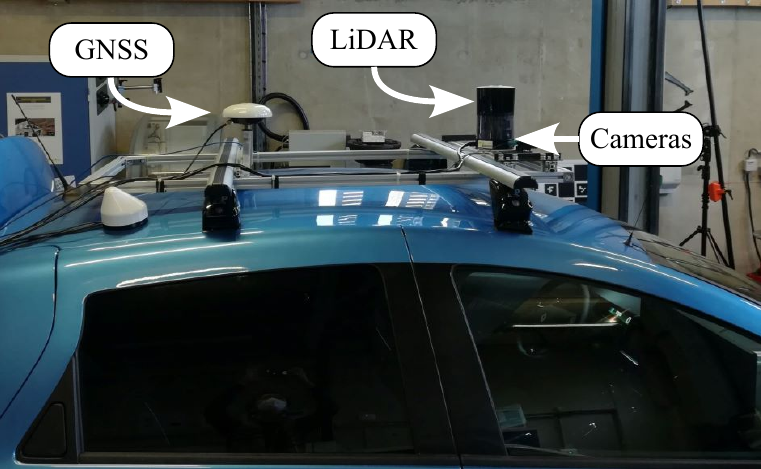}
 \caption{Roof of the experimental Renault ZOE vehicle equipped showing the GNSS antenna and the Hesai Pandora sensor combining a LiDAR with several cameras.}
 \label{fig:zoe-frame}
\end{figure}

\begin{figure}[t!]
 \centering
 \includegraphics[width=\columnwidth]{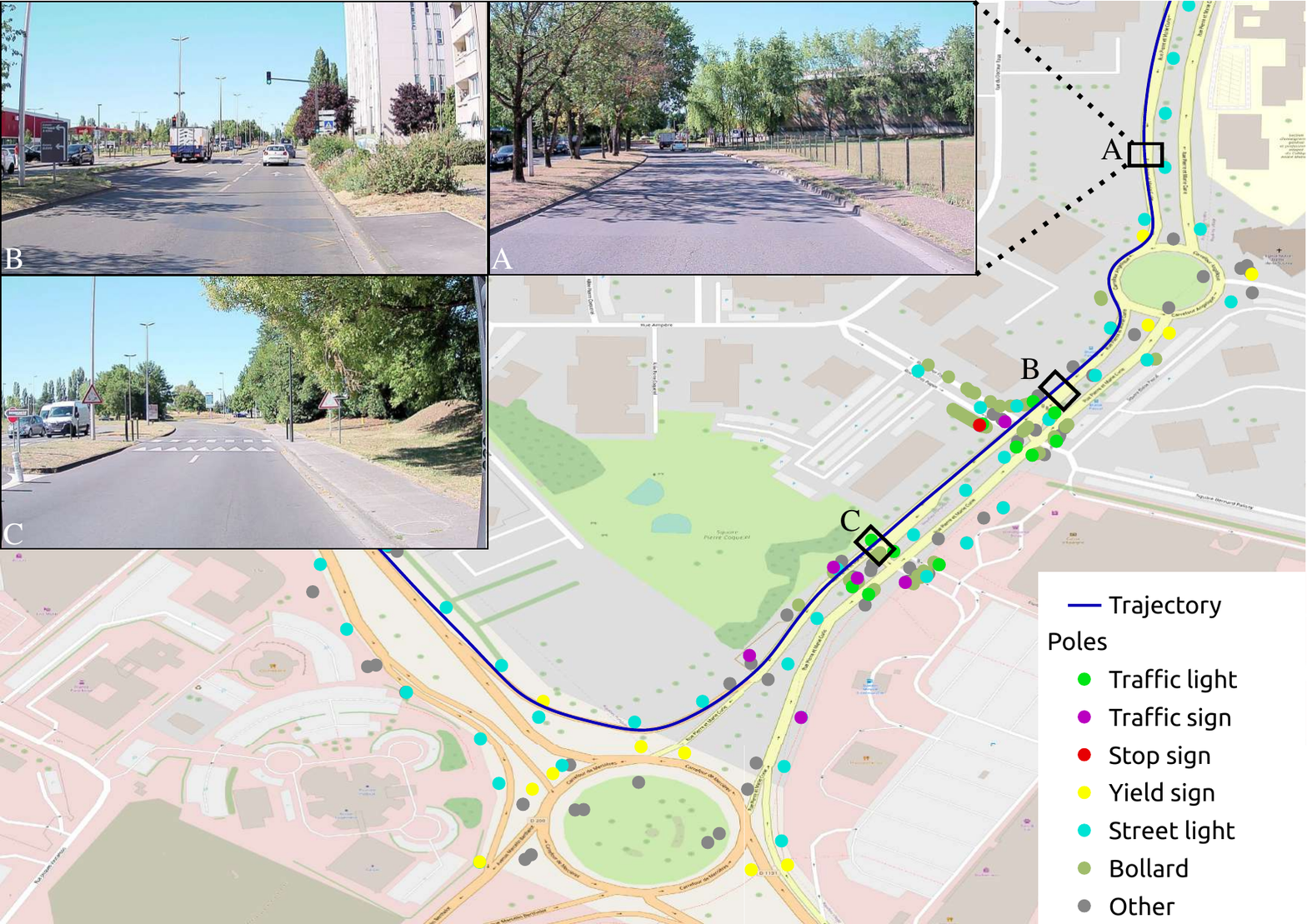}
 \caption{Experimental trajectory. The mapped pole-like features are displayed. The pictures illustrate the experimental conditions.}
 \label{fig:short-seq}
\end{figure}

The experiments were conducted with a Renault Zoe experimental vehicle equipped with several sensors: 
\begin{itemize}
    \item Wheel speed sensors [100 Hz]
    \item Septentrio mosaic X5 GNSS receiver with an automotive grade IMU [1 Hz]
    \item Hesai Pandora sensor combining a 40-layer LiDAR and 5 monocular cameras (4 grayscale cameras with a horizontal FOV of $129^\circ$ and one front RGB camera with a vertical FOV of $52^\circ$). The front and back grayscale cameras were not used in this study [10 Hz]
    \item Novatel SPAN-CPT GNSS/IMU with post-processed PPK computations for localization ground truth [50 Hz]  
\end{itemize}
The combination of sensors tested are:
\begin{itemize}
    \item \textbf{GNSS+DR} only uses  receiver, wheel speeds sensors and yaw rate. 
    \item \textbf{Front} uses GNSS+DR and the bearing measurements obtained from Pandora front color camera.
    \item \textbf{Left/Right} uses GNSS+DR and the bearing measurements obtained from Pandora left and right grayscale cameras.
    \item \textbf{All cameras} uses GNSS+DR and the bearing measurements obtained from Pandora left and right grayscale cameras and front color camera.
    \item \textbf{LiDAR} uses GNSS+DR and the pole measurements obtained from Pandora LiDAR.
\end{itemize}
Fig.~\ref{fig:zoe-frame} shows the roof of the experimental vehicle with the Hesai Pandora sensor.
We evaluated our pole-based localization framework on a 600~m-long section visible in Fig. \ref{fig:short-seq} extracted from datasets covering the city of Compi\`egne, France.

\subsection{Results and discussion}
The table \ref{tab:metrics-small-seq} summarizes the Root Mean Square (RMS) errors obtained from all various combinations on multiple datasets.

It is worth noting that the datasets were carefully acquired ensuring similar weather and traffic conditions for the majority of sequences, with the exception of the 05-19 sequence, which experienced higher traffic density. Then, the performance gap of detection methods between sequences due to variation in driving conditions is minimized.

Different behaviors occurred during these sequences. 
Firstly, for the 05-10 and 07-06 sequences, the performances obtained with all cameras are similar to the LiDAR. 
Yet, on the 05-10 sequence, the Left/Right is the combination obtaining the best results due to Front degradation which also affects the all-camera combination. 
On the 05-19 and 06-28 sequences, the LiDAR reached better performance than any camera combination. 
This is probably due to wrong associations when using camera measurements. 
Finally, on the 05-24 sequence, using all cameras is better than the LiDAR due also to miss-associations when using LiDAR measurements.

Globally, the position obtained using only GNSS and DR sensors is deeply improved except when miss-associations with the LiDAR occurred on the 05-24 sequence. 
The achieved performance is comparable with the LiDAR performance and adding all the cameras together instead of using only the Front or the Left/Right cameras improves globally the localization performance.

\begin{table}[t!]

\caption{RMS obtained for several combination of sensors for multiple datasets acquired under similar weather and traffic conditions} 
\label{tab:metrics-small-seq}
\smallskip
\begin{tabularx}{\columnwidth}{cccccc}
 \toprule
 Date  & GNSS+DR  & Front & Left/Right & All cameras & LiDAR  \\
 \cmidrule(lr){2-6}
 05-10 & 1.13 & 0.82 & \textbf{0.40} & 0.46 & 0.53 \\ 
 05-19 & 2.43 & 1.11 & 2.29 & 0.96 & \textbf{0.39}\\ 
 05-24 & 2.32 & 0.92 &  1.03 & \textbf{0.70} & 3.16 \\ 
 06-28 & 2.29 & 1.13 & 0.88 & 0.73 & \textbf{0.36}\\ 
 07-06   &  1.95 & 0.48 & 0.72 & \textbf{0.39} & 0.44\\
 \bottomrule
\end{tabularx}

\end{table}

\begin{figure}[t!]
 \centering
    \begin{subfigure}[b]{.48\textwidth}
       \includegraphics[width=\textwidth]{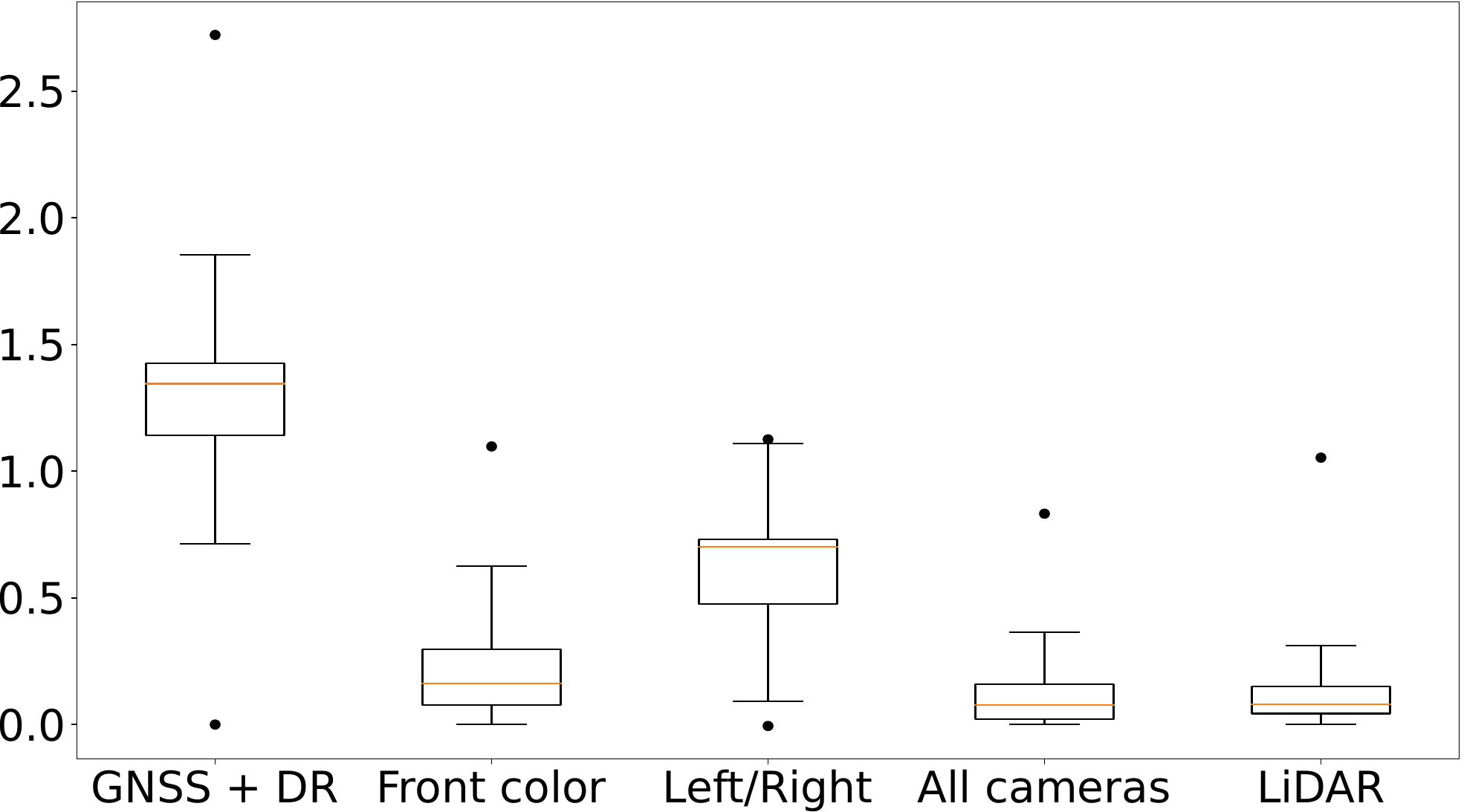}
       \caption{Cross-track errors.\medskip}
       \label{fig:bplot_lat_small_sequence}
    \end{subfigure}
    
    \begin{subfigure}[b]{.48\textwidth}
       \includegraphics[width=\textwidth]{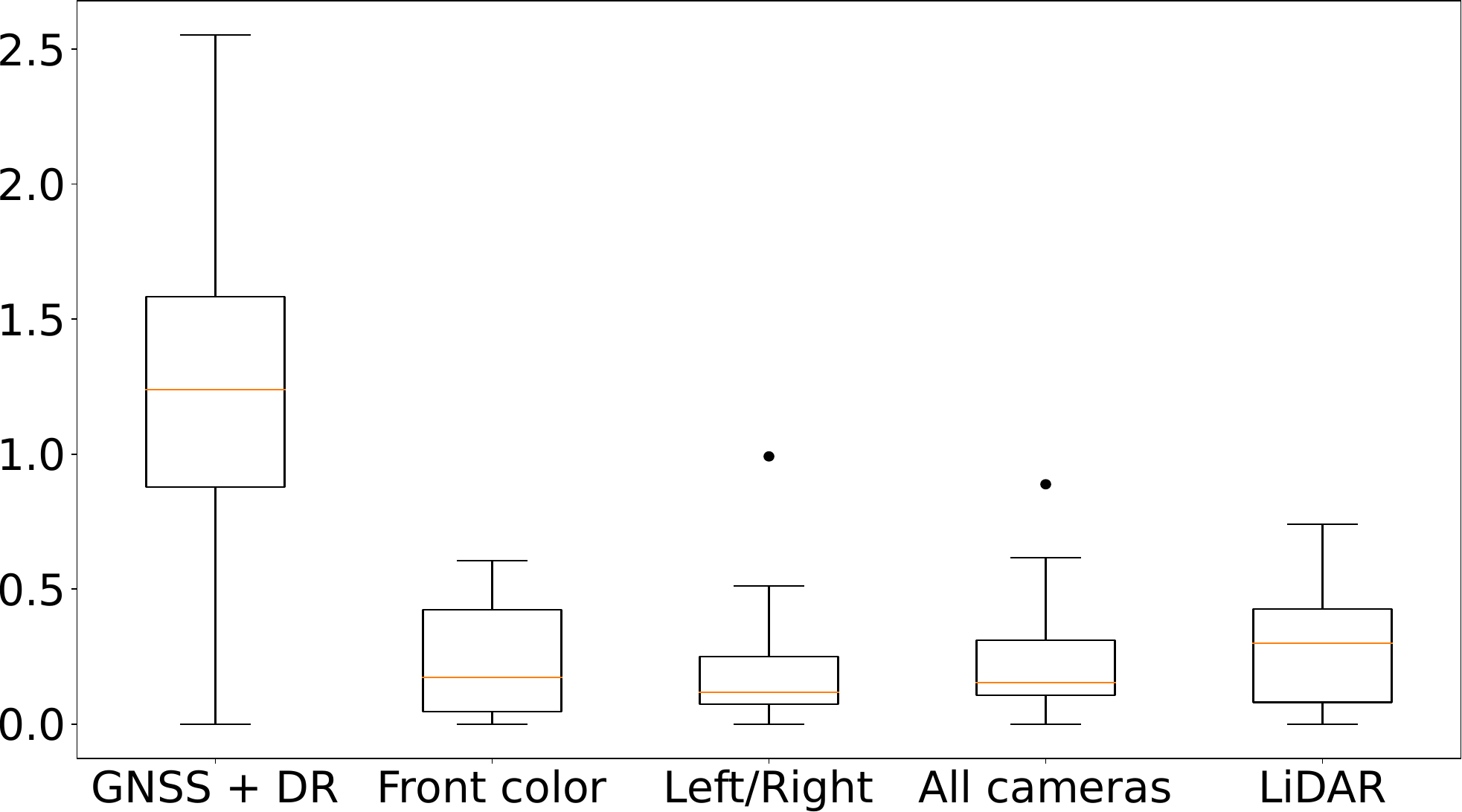}
       \caption{Along-track errors.}
       \label{fig:bplot_lon_small_sequence}
    \end{subfigure}
    \caption{Boxplots of the cross-track and along-track errors with different sensor setup obtained on the 07-06 sequence.}
\end{figure}

Each camera appears to contribute more to a specific component of the localization error.
In fact, when focusing on cross-track (CT) errors summarized in Fig. \ref{fig:bplot_lat_small_sequence} for the 07-06 sequence. 
All cameras performs better than LiDAR combination and this is, as expected, mainly due to the front color camera leading to an average error of less than \(40\) cm.

In terms of along-track (AT) errors, as visible in Fig. \ref{fig:bplot_lon_small_sequence}, even if Front is capable of improving AT accuracy, main improvement comes from the Left/Right combination, improving deeply all cameras solution, although extreme error values are still higher than LiDAR errors. 



\begin{figure}[t!]
 \centering
 \includegraphics[width=\columnwidth]{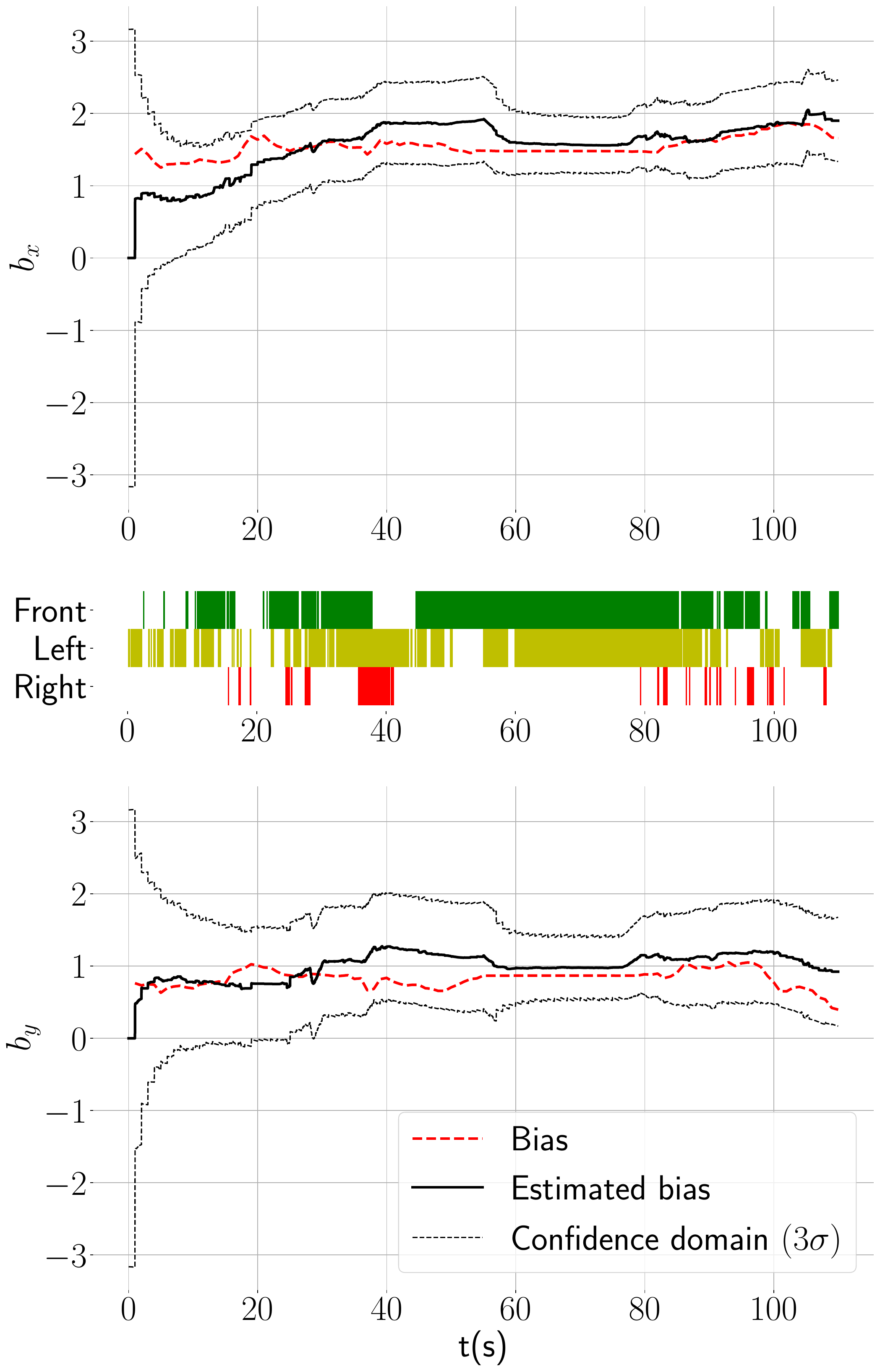}
 \caption{Biases in ENU frame obtained using the all cameras solution on the 07-06 sequence. Biases on $x$ and $y$ axes are respectively on top and bottom. The observations timestamps provided by the different cameras are summarized in the middle (Front camera in green, left camera in yellow and right camera in red).}
 \label{fig:biases}
\end{figure}

When focusing on biases estimation obtained during the 07-06 sequence using all cameras as shown in Fig. \ref{fig:biases}, the filter seems capable of estimating them even if some jumps are visible on the curves. Some of these jumps seems to be due to miss-associations between map features and detected poles. For example, a jump on \(b_y\) occurs around 30s after reception of right camera observations. A jump on \(b_x\) happens at the end of the sequence around 110s and seems to be correlated with reception of front camera observations.
Because the map-matching algorithm rely on an initial pose estimate, it is essential to correct the GNSS bias during the estimation.

Moreover, as shown in this figure, the results are primarily driven by the input from the front and left cameras. The right camera, detects fewer elements due to the majority of features being on the left side of the vehicle in this section. Removing the right camera would have had minimal impact on the overall solution.

Overall, our different detection methods face distinct challenges depending on the approach used, directly impacting localization accuracy. LiDAR may experience limitations in performance due to its simplistic detection process relying on multiple thresholding during a PCA procedure, which can result in a notable number of false positives and negatives.

On the other hand, cameras also encounter challenges in detection. These challenges primarily arise from the quality of training, particularly for the left and right cameras. The performance of the camera's detector is heavily influenced by the segmentation neural network used for image annotation, which is not optimal for wide-angle grayscale images. 

\section{Conclusion}
\label{sec:conclusion}
In this paper, we proposed to enhance a localization system based on GNSS and Dead Reckoning sensors by integrating pole-like feature detections and associations with a vector map. We proposed two different detection approaches using different sensors: one based on LiDAR sensor data where geometric filtering techniques are used and one based on object detection in camera images using automatically annotated data obtained from an image segmentation network. 

We compared the performance of LiDAR and multi-camera integration in terms of localization accuracy improvements on a complex peri-urban section containing multiple road features, mapped or not, and potential false positive detection sources. 

We showed that adding a front camera capable of detecting pole-like features can improve cross track positioning deeply.  This result was expected due to similarities with lane-marking based localization for lane-level positioning. Adding side cameras improves greatly the along-track positioning. Consequently, the combination of all these cameras provides localization performance similar to LiDAR integration. 

These results suggest that a multi-camera system is promising to replace or complete a LiDAR system, although further exploration is required to assess the robustness of the association process with the map, mitigate potential miss-associations and guarantee the integrity of the localization solution. 

In future work, our perception pipeline will be improved to enhance detection capabilities of the sensors and avoid false positives.  A comprehensive study will be undertaken to investigate the various factors influencing detection performance, including weather and traffic conditions, as well as the inherent characteristics of the detection methods themselves. Then, a particular attention will be given to the robustness of the data association and the estimation process. A study will be conducted to investigate the complementarity of LiDAR and cameras, the data association of multiple sources regarding the same features, and the benefits of such a system for localization.

\section*{Acknowledgment}
This work has been funded by the European project ERASMO~\cite{ERASMO23} (GSA/GRANT/03/2018) in the framework of the SIVALab laboratory between Renault and Heudiasyc.
\bibliographystyle{IEEEtran}
\bibliography{IEEEabrv,biblio}
\end{document}